\pdfoutput=1
\documentclass[11pt,a4paper]{article}
\usepackage[hyperref]{emnlp2020}

\usepackage{times}
\usepackage{latexsym}
\usepackage{todonotes}

\usepackage{microtype}

\aclfinalcopy 
\def\aclpaperid{767} 

\usepackage{multirow}
\usepackage{booktabs}
\usepackage[normalem]{ulem}
\usepackage{adjustbox}
\usepackage{xcolor}
\usepackage{pgfplots}
\usepgfplotslibrary{groupplots}
\pgfplotsset{compat=1.14}
\usepackage{eucal}
\usepackage{textcomp}
\usepackage{amsmath}

\newcommand{\fhalf}{F$_{0.5}$}
\newcommand{\pseudodata}{\textsc{GEC-pseudodata}~}

\usetikzlibrary{patterns}

\usepackage{xcolor}

\definecolor{seabornBlue}{RGB}{76,114,176}
\definecolor{seabornGreen}{RGB}{85,168,104}
\definecolor{seabornRed}{RGB}{196,78,82}

\definecolor{orangePumpkin}{RGB}{211,84,0}
\definecolor{orangeCarrot}{RGB}{230,126,34}
\definecolor{blueBelizeHole}{RGB}{41,128,185}
\definecolor{redAlizarin}{RGB}{231,76,60}
\definecolor{redNasturcianFlower}{RGB}{232,65,24}

\title{Grammatical Error Correction in Low Error Density Domains:\\ 
A New Benchmark and Analyses}

\author{Simon Flachs$^{1,2}$, Oph\'{e}lie Lacroix$^{3}$\thanks{~~Research conducted at Siteimprove.}, \\ \textbf{Helen Yannakoudakis$^{4}$, Marek Rei$^{5}$, Anders S{\o}gaard$^{2}$} \\
    $^{1}$ Siteimprove {\tt sfl@siteimprove.com} \\
    $^{2}$ University of Copenhagen {\tt soegaard@di.ku.dk}  \\
    $^{3}$ Alexandra Institute {\tt ophelie.lacroix@alexandra.dk} \\
    $^{4}$ King's College London {\tt helen.yannakoudakis@kcl.ac.uk} \\
    $^{5}$ Imperial College London {\tt marek.rei@imperial.ac.uk}\\
  }

\date{}

\begin{document}
\maketitle
\begin{abstract}

Evaluation of grammatical error correction (GEC) systems has primarily focused on essays written by non-native learners of English, which however is only part of the full spectrum of GEC applications. 
We aim to broaden the target domain of GEC and release CWEB, a new benchmark for GEC consisting of website text generated by English speakers of varying levels of proficiency. 
Website data is a common and important domain that contains far fewer grammatical errors than learner essays, which we show presents a challenge to state-of-the-art GEC systems.
We demonstrate that a factor behind this is the inability of systems to rely on a strong internal language model in low error density domains. 
We hope this work shall facilitate the development of open-domain GEC models that generalize to different topics and genres.

\end{abstract}

\section{Introduction}

Grammatical error correction (GEC) is the task of automatically editing text to remove grammatical errors; for example: [\textit{A link to registration can also be found \sout{at} \textbf{on} the same page.}]. 
GEC systems so far have primarily focused on correcting essays produced by English-as-a-second-language (ESL) learners, providing fast and inexpensive feedback to facilitate language learning. 
However, this is only one target domain in the full spectrum of GEC applications. GEC models can also help to improve written communication outside of the formal education setting. 
Today the largest medium of written communication is the internet, with approximately $380$ new websites created every minute.\footnote{\url{https://www.millforbusiness.com/how-many-websites-are-there/}}
Ensuring grammatical correctness of websites helps facilitate clear communication and a professional commercial presentation. 
Therefore, it is important that GEC models perform well in the open-domain setting and generalize, not only to writing produced in the educational context, but also to language production ``in the wild''. Website data specifically represent a broad and diverse range of writing and constitute a major part of what people read and write on an everyday basis. 

\begin{figure}[!t]
\scalebox{0.92}{
\begin{tikzpicture}
        \begin{axis}[
            symbolic x coords={
            JFLEG, FCE, CoNLL14, GMEG Wiki, GMEG Yahoo, W\&I-A, W\&I-B, W\&I-C, LOCNESS, AESW, CWEB-G, CWEB-S},
            xtick=data,
            x tick label style={rotate=45, anchor=east},
            ylabel=\% erroneous tokens,
          ]
            \addplot[ybar,fill=seabornBlue,postaction={pattern=north east lines}] coordinates {
                
                (JFLEG, 19.064595834804095)
                (FCE, 13.180864256415148)
                (CoNLL14, 11.340565286624203)
                (GMEG Wiki, 8.010833022039597)
                (GMEG Yahoo, 7.406697718294912)
                (W\&I-A, 18.992767211358156)
                (W\&I-B, 12.764790277660563)
                (W\&I-C, 6.0915838117910655)
                (LOCNESS, 4.65458320716356)
                (AESW, 2.7005337703420538)
                (CWEB-G, 0)
                (CWEB-S, 0)
            };
                        \addplot[ybar,fill=seabornGreen] coordinates {

                (CWEB-G, 2.357507812887544)
                (CWEB-S, 1.7098263868591805)
            };
        \end{axis}
    \end{tikzpicture}
}

\caption{Percentage of erroneous tokens per domain. CWEB-G/S are our newly devised datasets.}
\label{fig-err_tokens}
\end{figure}
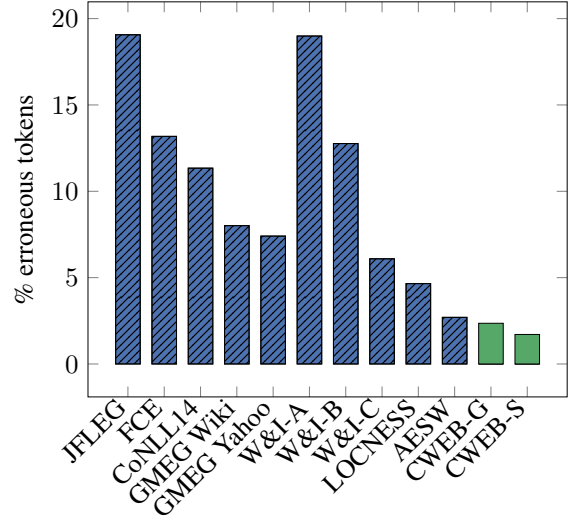
\begin{table*}[th]
    \centering
    \begin{small}
    \begin{tabular}{c|p{0.7\textwidth}}
        \toprule
        \textbf{Error type} & \textbf{Example sentence} \\
        \midrule
        \textsc{Verb:Sva} & They develop positive relationships with swimmers and members, and \textcolor{red}{\sout{promotes}} \textbf{promote} programs in order to generate more participation. \\
        \textsc{Morph} / \textsc{Orth} & In a small \textcolor{red}{\sout{agriculture}} \textbf{agricultural} town on the east side of Washington \textcolor{red}{\sout{state}} \textbf{State}  called Yakima. \\
        \textsc{Prep} & 
         [\dots] the distance between the two should be \textcolor{red}{\sout{on}} \textbf{of} the order of 50 microns. \\
        \bottomrule
    \end{tabular}
    \end{small}
    \caption{Example sentences from the CWEB dataset. Erroneous text is struck through and corrections are in bold.}
    \label{tab-example}
\end{table*}

This work highlights two major prevailing challenges of current approaches to GEC: 
\emph{domain adaptation} and \emph{low precision} in texts with low error density. 
Previous work has primarily targeted essay-style text with high error density (see Figure \ref{fig-err_tokens});
however, this lack of diversity means that it is not clear how systems perform on other domains and under different error distributions \citep{sakaguchi2017gec}.\footnote{\citet{leacock2010automated} highlighted the variations in the distribution of errors in non-native and native English writings.}

Current publicly available datasets are restricted to non-native English essays [e.g. FCE \cite{yannakoudakis2011fce}; 
CoNLL14 \citep{Ng2014conll}], 
student essays [W\&I+LOCNESS \cite{bea2019,granger1998computer}] or target a specific domain [scientific writing; AESW \cite{daudaravicius2016report}].
Supervised systems trained on specific domains are less likely to be as effective at correcting distinctive errors from other domains, as is the case for systems trained on learner data with different native languages  \citep{chollampatt2016adapting,nadejde2019personalizing}.
The recent BEA 2019 shared task \citep{bea2019} encouraged research in the use of low-resource and unsupervised approaches; however, evaluation primarily targeted the restricted domain of student essays. 
We show that when applied to data outside of the language
learning domain, current state-of-the-art systems exhibit low precision due to a  tendency to over-predict errors.
Recent work tackled the domain adaptation problem, and released GEC benchmarks from Wikipedia data and online comments [GMEG Wiki+Yahoo \cite{napoles2019enabling}]. However, these datasets present a high density of errors and represent a limited subset of the full distribution of errors in online writing. 
 \\ \\ \textbf{Contributions:} We (i) release a new dataset, CWEB (\textbf{C}orrected \textbf{Web}sites), of website data that is corrected for grammatical errors;\footnote{\url{https://github.com/SimonHFL/CWEB}} (ii) systematically compare it to previously released GEC corpora; (iii) benchmark current state-of-the-art GEC approaches on this data and demonstrate that they are heavily biased towards existing datasets with high error density, 
even after fine-tuning on our target domain; (iv) perform an analysis showing that a factor behind the performance drop is the inability of systems to rely on a strong internal language model in low error density domains. 
\\ \hspace*{2mm} We hope that the new dataset will contribute towards the development of robust GEC models in the open-domain setting.

\section{CWEB Dataset}

\begin{table}[t]
\centering
\begin{small}
\begin{tabular}{@{~~}l@{~~}l|@{~~}r@{~~}r|@{~~}r@{~~}}
    \toprule
     & & \textbf{CWEB-S} & \textbf{CWEB-G} & \textbf{Total} \\
    \midrule
    \parbox[t]{3mm}{\multirow{3}{*}{\rotatebox[origin=c]{90}{\textbf{Dev}}}} & sent. & 2,862 & 3,867 & 6,729 \\
     & tokens & 68,857 & 79,689 & 148,546 \\
     & edits  & 895 & 1595 & 2490 \\
    \midrule
    \parbox[t]{3mm}{\multirow{3}{*}{\rotatebox[origin=c]{90}{\textbf{Test}}}} & sent. & 2,864 & 3,981 & 6,845 \\
     & tokens & 68,459 & 80,684 & 149,143 \\
     & edits  & 1004 & 1679 & 2683 \\
    \midrule
    \midrule
    \parbox[t]{3mm}{\multirow{4}{*}{\rotatebox[origin=c]{90}{\textbf{Total}}}} & sent. & 5,726 & 7,848 & 13,574 \\
     & tokens & 137,316 & 160,373 & 297,689 \\
    \cmidrule{2-5}
     & websites & 453 & 625 & 1,078 \\
     & parag. & 659 & 630 & 1,289 \\
    \bottomrule
\end{tabular}
\end{small}
\caption{Distribution of sentences and tokens in the CWEB dataset.}
\label{tab-data}
\end{table}

\begin{table*}[th]
    \centering
    \begin{small}
    \begin{tabular}{lrccccccccc}
        \toprule
         & \textbf{\# sents} & \textbf{type-token} & \textbf{tok/sent} & \textbf{err. sents} (\%) & \textbf{edits/sent} & \textbf{\# annotators} & \textbf{sent-$\mathcal{K}$} & \textbf{NEs/sents}  \\
        \midrule
        JFLEG & 747 & 0.44 & 18.9 & 86.4 & 3.6 & 4 & 0.53 & 0.35 \\
        \midrule
        FCE & 2,695 & 0.39  & 15.6 & 67.8 & 2.6 & 1 & -$^\dagger$ & 0.59 \\
        \midrule
        CoNLL14 & 1,312 & 0.39 & 22.9 & 75.8 & 2.7 & 2 & 0.25 & 0.31 \\
        \midrule
        W\&I-A & 1,036 & 0.43 & 18.0 & 80.5 & 3.6 & 1 & -$^\dagger$ & 0.58 \\
        W\&I-B & 1,285 & 0.45 & 18.4 & 72.1 & 2.7 & 1 & -$^\dagger$ & 0.52 \\
        W\&I-C & 1,068 & 0.47 & 20.1 & 53.8 & 1.9 & 1 & -$^\dagger$ & 0.78 \\
        \midrule
        LOCNESS & 988 & 0.47 & 23.4 & 52.2 & 1.8 & 1 & -$^\dagger$ & 0.77 \\
        \midrule
        GMEG wiki & 992 & 0.55 & 26.9 & 82.3 & 2.5 & 4 & 0.43 & 2.83 \\
        GMEG yahoo & 1,000 & 0.46 & 16.9 & 50.5 & 2.7 & 4 & 0.51 & 0.59 \\
        \midrule
        AESW & 52,124 & 0.52 & 23.9 & 36.1 & 1.6  & 1 & -$^\dagger$ & 0.93 \\
        \midrule
        CWEB-S & 2,864 & 0.56 & 23.9 & 24.5 & 1.5 & 2 & 0.39 & 1.44 \\
        CWEB-G & 3,981 & 0.53 & 20.3 & 25.6 & 1.9 & 2 & 0.44 & 1.04 \\
        \bottomrule
    \end{tabular}
    \end{small}
    \caption{Statistics on GEC Corpora; type--token is the average ratio of vocabulary size by the total number of tokens (calculated as an average over a sliding window of $1,000$ tokens); ratio of edits per sentence is calculated on erroneous sentences; sent-$\mathcal{K}$ is sentence-level Cohen's Kappa score ($\dagger$: calculated for datasets with $>1$ annotator); NEs stands for Named Entities (extracted using Spacy).
    }
    \label{tab-stats}
\end{table*}

We create a new dataset of English texts from randomly sampled websites, and annotate it for grammatical errors. 
The source texts are randomly selected from the first $18$ dumps of the CommonCrawl\footnote{\url{https://commoncrawl.org/}} dataset and represent a wide range of data seen online such as blogs, magazines, corporate or educational websites. 
These include texts written by native or non-native English speakers and professional as well as amateur online writers. 

\paragraph{Text Extraction}
To ensure English content, we exclude websites with country-code top-level domains; e.g., .fr, .de. 
We use the \texttt{jusText}\footnote{\url{https://github.com/miso-belica/jusText}} tool to retrieve the content from HTML pages (removing boilerplate elements and splitting the content into paragraphs).
We heavily filter the data by removing paragraphs which contain non-English\footnote{Using the
\href{https://github.com/Mimino666/langdetect}{\texttt{langdetect}}
 package.} and incomplete sentences. 
To ensure diversity of the data, we also remove duplicate sentences.
Among the million sentences gathered, we select paragraphs randomly.

We split the data with respect to where they come from: sponsored\footnote{top-level domains: .gov, .edu, .mil, .int, and .museum.} (CWEB-S) or generic\footnote{top-level domains: .com, .info, .net, .org.} (CWEB-G) websites. 
The sponsored data represent a more focused domain (professional writing) than the generic one which includes writing from various proficiency levels.

\paragraph{Annotation}
The data is corrected for errors by two expert annotators, trained for correcting grammatical errors in English text: not attempting to rewrite the text nor make fluency edits, but rather to make minimal edits -- minimum number of edits to make the text grammatical.
During error annotation, the annotators have access to the entire paragraph in which a sentence belongs, therefore using the context of a sentence to help them in the correction. 
Examples of erroneous sentences from our data are shown in Table~\ref{tab-example}. 
Annotator agreement is calculated at the sentence level using Cohen's Kappa, i.e. we calculate whether annotators agree on which sentences are erroneous. This approach is preferable to relying on exact matching of error corrections, as 
as there are often many different ways to correct a sentence \citep{bryant2015far}.
Kappa is $0.39$ and $0.44$ for sponsored (CWEB-S) and generic website (CWEB-G) data respectively, and Table~\ref{tab-stats} presents how our agreement results compare to those of existing GEC datasets. The table also includes a number of other statistics, and the different datasets are further analyzed, compared and contrasted in Section \ref{sec:analysis}. 

The texts are tokenized using SpaCy\footnote{\url{https://spacy.io/}} and automatically labeled for error types (and converted into the M2 format) 
using the ERRor ANnotation Toolkit (ERRANT) \citep{bryant2017automatic}.

\paragraph{Release}
For each dataset, we release a development and a test set: we propose a roughly equal division of the data into the two splits, which presents a fair amount of errors to evaluate on (see Table~\ref{tab-data}).

To avoid copyright restrictions, we split the collected paragraphs into sentences and shuffle all sentences in order to break the original and coherent structure that would be needed to reproduce the copyrighted material. This approach has successfully been used in previous work for devising web-based corpora \citep{Schaefer2015, biemann2007leipzig}.
The data is available at \url{https://github.com/SimonHFL/CWEB}.

\begin{table*}[th]
    \centering
    \begin{small}
    \begin{tabular}{l@{~~}r@{~~~}r@{~~~}r@{~~~}r@{~~~}r@{~~~}r@{~~~~}rrr@{~~~}rr@{~~~}r}
        \toprule & \multirow{2}{*}{\textbf{JFLEG}} & \multirow{2}{*}{\textbf{FCE 2.1}} & \multirow{2}{*}{\textbf{CoNLL14}} & \multicolumn{3}{c}{\textbf{W\&I}} & \multirow{2}{*}{\textbf{LOCNESS}} &  \multicolumn{2}{c}{\textbf{GMEG}} & 
        \multirow{2}{*}{\textbf{AESW}} &  \multicolumn{2}{c}{\textbf{CWEB}} \\
         \cmidrule(r){5-7} \cmidrule(r){9-10} \cmidrule(r){12-13} & & & & \textbf{A} & \textbf{B} & \textbf{C} & & \textbf{Wiki} & \textbf{Yahoo} & &  \textbf{G} & \textbf{S} \\
        \midrule


        \textsc{Punct} & 147.7 & 112.3 & 65.5 & 244.8 & 188.2 & 100.4 & 152.3 & 230.0 & 194.0 & 80.6 & 48.9 & 48.7\\
        \textsc{Verb} & 233.5 & 176.7 & 200.5 & 300.0 & 202.5 & 79.4 & 19.9 & 48.1 & 24.2 & 17.8 & 23.4 & 13.1\\
        \textsc{Other} & 295.6 & 138.3 & 158.1 & 237.3 & 136.7 & 57.4 & 43.3 & 93.8 & 98.0 & 42.7 & 31.6 & 21.0\\
        \textsc{Det} & 180.7 & 149.1 & 134.9 & 159.1 & 124.1 & 65.8 & 16.4 & 40.3 & 22.6 & 33.7 & 20.9 & 19.7\\
        \textsc{Noun} & 167.7 & 105.4 & 116.8 & 139.8 & 89.0 & 49.9 & 32.4 & 63.6 & 26.2 & 16.8 & 19.6 & 12.8\\
        \textsc{Prep} & 107.1 & 113.8 & 92.7 & 137.2 & 114.4 & 64.9 & 28.1 & 37.1 & 21.1 & 11.4 & 15.6 & 9.8\\
        \textsc{Spell} & 242.5 & 107.8 & 26.0 & 79.3 & 36.3 & 16.3 & 51.0 & 86.9 & 68.0 & 5.1 & 3.8 & 2.4\\
        \midrule
        \textsc{All} & 1675.6 & 1084.9 & 919.6 & 1561.2 & 1050.7 & 504.1 & 400.6 & 732.3 & 635.3 & 239.2 & 208.9 & 147.2\\
        \bottomrule
    \end{tabular}
    \end{small}
    \caption{Number of error occurrences for the most frequent error types (per $10,000$ token).}
    \label{tab-err-frequency}
\end{table*}


\section{GEC Corpora}

We compare our data with existing GEC corpora which cover a range of domains and proficiency levels. 
Table~\ref{tab-stats} presents a number of different statistics and Table~\ref{tab-err-frequency} their error-type frequencies.\footnote{See links to downloadable versions in Appendix \ref{app-datasets}}

\subsection{English as a second language (ESL)}

\paragraph{JFLEG} \cite{napoles2017jfleg} The JHU Fluency-Extended GUG corpus consists of sentences written by English language learners (with different proficiency levels and L1s) for the TOEFL\textregistered~ exam, covering a range of topics.
Texts have been corrected for grammatical errors and fluency.

\paragraph{FCE} \cite{yannakoudakis2011fce} consists of $1,244$ error corrected texts produced by learners taking the First Certificate in English exam, which assesses English at an upper-intermediate level.
We use the data split made available for the BEA GEC shared task 2019 \cite{bea2019}.

\paragraph{CoNLL14} \cite{Ng2014conll}
consists of (mostly argumentative) essays written by ESL learners from the National University of Singapore, which are annotated for grammatical errors by two native speakers of English.

\paragraph{Write\&Improve (W\&I)} \cite{bea2019} Cambridge English Write \& Improve \cite{Yannakoudakis2018} is an online web platform that automatically provides diagnostic feedback to non-native English-language learners, including an overall language proficiency score based on the Common European Framework of Reference for Languages (CEFR).\footnote{\url{https://www.cambridgeenglish.org/exams-and-tests/cefr/}} The W\&I corpus contains $3,600$ texts across $3$ different CEFR levels -- A (beginner), B (intermediate), and C (advanced) -- that have been annotated for errors.\footnote{\label{note1}Since error corrections on test sets are not publicly available, we carry out our analyses on the development sets.}

\subsection{Other Corpora}

\paragraph{LOCNESS} \cite{bea2019,granger1998computer} The LOCNESS corpus consists of essays written by native English students. A sample of $100$ essays has been annotated for errors with a 50:50 development/test split.\footnote{See footnote \ref{note1}.}

\paragraph{GMEG Wiki}
\cite{napoles2019enabling} is devised based on edits in the Wikipedia revision history, and the writing therefore represents formal articles. 
Note that collecting sentences based on edits in the Wikipedia revision history introduces a substantial bias.\footnote{Sentences that have been edited are more likely to contain grammatical errors, and grammatical errors will therefore be over-represented. This is reflected in the 82.3\% erroneous sentence rate (see Table~\ref{tab-stats}).} This means that evaluation results on this benchmark are not truly representative of how a system would perform when applied to realistic online data and full-length articles.

\paragraph{GMEG Yahoo} \cite{napoles2019enabling} comprises paragraphs of informal web posts gathered from answers in the \textit{Yahoo! Answers} platform. The style is informal, and contains slang terms and non-conventional mechanics. 

\paragraph{AESW} \cite{daudaravicius2016report} was released as part of the Automated Evaluation of Scientific Writing Shared Task. It is a collection of text extracts from published journal articles (mostly in physics and mathematics) along with their (sentence-aligned) corrected counterparts.\footnote{We exclude sentences that use AESW's normalization scheme (e.g. citations replaced with \_\_CITE\_\_), as the models we use are not trained with these special tokens.}

\begin{table*}[t]
    \centering
    \begin{small}
    \begin{tabular}{ll@{~~}r@{~}r@{~~~}rrr@{~~~}r@{~~~}r@{~~~}r@{~~~~}r@{~~~}rr@{~~~}r}
        \toprule & \multirow{2}{*}{\textbf{JFLEG}} & \multirow{2}{*}{\textbf{FCE}} & \multirow{2}{*}{\textbf{CoNLL14}} & \multicolumn{3}{c}{\textbf{W\&I}} & \multirow{2}{*}{\textbf{LOCNESS}} &  \multicolumn{2}{c}{\textbf{GMEG}} & 
        \multirow{2}{*}{\textbf{AESW}} &  \multicolumn{3}{c}{\textbf{CWEB}} \\
         \cmidrule(r){5-7} \cmidrule(r){9-10} \cmidrule(r){12-14} 
        & & & & \textbf{A} & \textbf{B} & \textbf{C} & & \textbf{Wiki} & \textbf{Yahoo} & &  \textbf{G} & \textbf{S} & \textbf{G+S} \\
        \midrule \midrule
        & \multicolumn{13}{c}{\textsc{GEC-pseudodata} system} \\
        \midrule
        P & 55.73 & 55.11 & 44.96 & 54.89 & 54.86 & 44.53 & 47.09 & 52.81 & 37.57 & 14.05 & 21.34 & 17.27 & 19.97 \\
        R & 38.73 & 41.61 & 29.03 & 37.92 & 35.14 & 32.04 & 34.13 & 23.02 & 32.26 & 13.24 & 23.00 & 15.75 & 20.28 \\
        F$_{0.5}$ & 51.13 & 51.75 & 40.35 & 50.38 & 49.32 & 41.31 & 43.77 & 41.89 & 36.00 & 13.88 & 21.58 & 16.91 & 19.98 \\ 
        \midrule \midrule
        & \multicolumn{13}{c}{\textsc{PIE} system} \\
        \midrule
        P & 51.04 & 49.55 & 43.47 & 50.24 & 49.12 & 39.12 & 32.77 & 44.71 & 33.08 & 8.78 & 14.29 & 5.73 & 10.80 \\ 
        R & 35.21 & 36.34 & 27.93 & 36.10 & 31.20 & 27.13 & 23.11 & 19.66 & 26.97 & 9.67 & 18.91 & 8.78 & 15.11 \\
        F$_{0.5}$ & 46.74 & 46.19 & 38.95 & 46.59 & 44.06 & 35.94 & 30.24 & 35.58 & 31.29 & 8.94 & 14.98 & 6.15 & 11.43 \\
        \bottomrule
    \end{tabular}
    \end{small}
    \caption{Scores of two \textsc{sota} GEC systems on each domain. For both systems performance is substantially lower on CWEB than ESL domains. 
    Scores are calculated against each individual annotator and averaged
    }
    \label{tab-results}
\end{table*}

\section{System Performance}

We evaluate performance on GEC benchmarks for two approaches to GEC that currently have state-of-the-art performance on CoNLL14. 
The first approach, that we refer to as \pseudodata and is proposed by \citet{kiyono2019empirical},\footnote{\url{www.github.com/butsugiri/gec-pseudodata};
We use the PRETLARGE+SSE (finetuned) model.} uses a transformer-based seq2seq model.  
The second approach uses the PIE system \cite{awasthi2019parallel}\footnote{\url{www.github.com/awasthiabhijeet/PIE}} which leverages a BERT-based architecture for local sequence transduction tasks.
Both models are pre-trained on synthetic errors and fine-tuned on learner data from the train section of FCE \citep{yannakoudakis2011fce}, Lang-8 \citep{mizumoto2011lang8}, and NUCLE \citep{dahlmeier2013building} and for GEC-PSEUDODATA additionally on the W\&I train split \citep{bea2019}.

Performance is evaluated using the $F_{0.5}$ metric calculated by ERRANT \citep{bryant2017automatic}.\footnote{\url{www.github.com/chrisjbryant/errant}} However, the more annotators a dataset has, the higher score a system will get on this data \citep{bryant2015far}. In order to perform a fair comparison of systems across datasets with a different number of annotators, we calculate the ERRANT score against each individual annotator and then take the average to get the final score.

Evaluation results are presented in Table \ref{tab-results}.
Across all datasets, we observe lower scores with the PIE system  ($-6.05$ \fhalf~on average), while \pseudodata is consistently better. Overall $F_{0.5}$ ranges from around $30$ to $52$ for most datasets; however, when the models are evaluated on CWEB and AESW, we observe a substantial drop in performance, with the lowest $F_{0.5}$ score being the PIE system on CWEB-S ($6.15$).
Precision, in particular, suffers due to the systems over-correcting sentences that should remain unchanged.

Using the \pseudodata system, on average, we find a higher \fhalf~on ESL corpora (JFLEG, FCE, CoNLL, W\&I) compared to non-ESL ones ($47.4$ vs.~$29.0$). 
This demonstrates that GEC systems trained on language learning data do not perform as well on other domains and further work is needed to improve their generalization.

\subsection{Fine-tuning}

We investigate the extent to which the \pseudodata system can be adapted to our domain, and fine-tune it using our development sets.\footnote{We use the fine-tuning parameters of \citet{kiyono2019empirical}.} We take $1,000$ sentences from each of the development sets of CWEB-G and CWEB-S and use them as a development set for this experiment. The remaining $4,729$ sentences of our development sets are used as training data for fine-tuning the GEC system. 

In Table \ref{tab-finetuning}, we can see that fine-tuning substantially improves performance (around $+10.0$ \fhalf~across all CWEB sets). In particular, precision is improved ($+20.8$/$+18.6$ on CWEB-G/S) 
at the expense of recall ($-6.4$/$-2.8$ on CWEB-G/S). 
However, performance is still low compared to the language learning domain (\fhalf~of at least $41$),
further indicating that there is scope for developing more robust and general-purpose, open-domain GEC systems.
For the purpose of future benchmarking, Appendix \ref{non-averaged-scores} lists the system\textquotesingle s ERRANT scores based on both annotators -- as opposed to the average of individual annotator scores reported in Table \ref{tab-finetuning}.

\begin{table}[t]
    \centering
    \begin{small}
    \begin{tabular}{lrcc}
        \toprule
        & \textbf{P} & \textbf{R} & \textbf{\fhalf}  \\
        \midrule
        CWEB-G & 42.09 & 16.56 & 32.01 \\
            CWEB-S & 35.91 & 12.96 & 26.46 \\
            \textbf{{CWEB}} (G+S) & 39.89 & 15.2 & 30.0 \\
        \bottomrule
    \end{tabular}
    \end{small}
    \caption{Scores of the \pseudodata system fine-tuned on CWEB data. 
    Fine-tuning yields substantial improvements, but scores are still worse than on ESL domains. 
    Scores are calculated against each individual annotator and averaged.
    }
    \label{tab-finetuning}
\end{table}


\section{Analysis}
\label{sec:analysis}

In order to assess the impact our new dataset can have on the GEC field, we carry out analyses to show 1) to what degree the domain of our data is different from existing GEC corpora, and how existing GEC systems are affected by the domain shift; and 2) that a factor behind the performance drop on CWEB data is the inability of systems to rely on a strong internal language model in low error density domains.

\subsection{Domain Shift}

Moving from error correction in learner texts to error correction in diverse, online texts, many of which are written by professional writers, amounts to a drift in data distribution. In general, distributional drift comes in different flavors; 
given two distributions $P(\mathbf{X},\mathbf{Y})$ and $Q(\mathbf{X},\mathbf{Y})$: 

\vspace{-0.1cm}
\paragraph{Covariate shift} concerns change in the marginal distribution of the independent variable, i.e., $P(\mathbf{X})\neq Q(\mathbf{X})$. 
In the context of grammatical errors, this refers to the degree to which the type of sentences written varies between domains. Table~\ref{tab-stats} clearly shows covariate shift effects: see, for example, differences in vocabulary variation (measured by the type--token ratio) and the frequency of named entities.

\vspace{-0.1cm}
\paragraph{Label bias} describes the change in distribution of the dependent variable, i.e., $P(\mathbf{Y})\neq Q(\mathbf{Y})$. In terms of GEC, this refers to the difference in error distributions across domains. 
In Table \ref{tab-stats}, we can see that CWEB data contains errors that are substantially more sparse than other domains -- a smaller proportion of sentences are erroneous, and these erroneous sentences also contain fewer edits compared to other domains. Additionally, looking at Table \ref{tab-err-frequency}, we can see that almost all error types are substantially less frequent in our data than in existing benchmarks -- for example, spelling errors are $38$ times more prevalent in GMEG Wiki compared to CWEB-S.

Moving from learner text to web data involves both forms of drift: covariate shift and label bias. 
We further analyze the effects of these shifts on system performance.

\subsubsection{Impact of Error Density}\label{subsec-error-prop}

To demonstrate that the error density of corpora has a substantial impact on the performance of GEC systems, 
we vary the proportion of erroneous sentences in each dataset by either removing correct sentences or by adding correct sentences of the same domain.\footnote{For each dataset, we apply the gold corrections on incorrect sentences, creating new examples of in-domain, correct sentences, 
which are then randomly selected for inclusion.} 
By fixing the frequency of errors across datasets, we can observe, in isolation, how the systems are affected by co-variate shift across domains.
Precision as a function of the proportion of erroneous sentences for selected datasets\footnote{Scores for all datasets can be found in Appendix~\ref{app-percentage}.} is presented in Figure~\ref{fig-percentage} (recall is unchanged).

For each domain, we observe precision being highly sensitive to the proportion of errors. This indicates that differences in error distribution across domains (i.e. label bias) is likely to be a large contributor to performance drop.
We also observe the effect of covariate shift across the datasets: 
while the percentage of erroneous sentences is the same, precision differs for the different datasets which suggests that covariate shift across domains has an impact on the performance of the system. 

\begin{figure}[t]
    \centering
    \begin{tikzpicture}
\begin{axis}[
	width=\linewidth, 
  ylabel={precision},
  xlabel={proportion of erroneous sentences},
  legend style = {at={(1.02,1.25)}, legend columns=2},
  legend = {},
  ymax=70,
  ymin=0
  ]
  \addplot[smooth, mark=square*, dashed, line width=0.6pt, color=seabornBlue] coordinates { 
  (0.1, 15.35) (0.2, 26.95) (0.3, 35.6) (0.4, 42.78) (0.5, 48.9) (0.6, 53.91) (0.7, 57.03) (0.8, 58.86) (0.9, 60.65) (1.0, 61.87)
  };\addlegendentry{FCE (PSEUDO)}
  \addplot[smooth, mark=square, line width=0.6pt, color=seabornGreen] coordinates { 
  (0.1, 20.67) (0.2, 31.94) (0.3, 38.96) (0.4, 43.18) (0.5, 46.9) (0.6, 49.58) (0.7, 51.16) (0.8, 52.16) (0.9, 52.72) (1.0, 53.49)
  };\addlegendentry{FCE (PIE)}
  \addplot[smooth, mark=*,  dashed, line width=0.6pt, color=orangePumpkin] coordinates { (0.1, 8.62) (0.2, 15.855) (0.3, 22.755) (0.4, 28.635) (0.5, 33.435) (0.6, 38.18) (0.7, 42.357) (0.8, 47.715) (0.9, 50.907) (1.0, 53.012)   };\addlegendentry{Wiki (PSEUDO)}
  \addplot[smooth, mark=otimes, line width=0.6pt, color=orangeCarrot] coordinates { (0.1, 6.325) (0.2, 12.07) (0.3, 17.655) (0.4, 21.605) (0.5, 26.433) (0.6, 30.857) (0.7, 35.987) (0.8, 39.188) (0.9, 42.633) (1.0, 44.877) };\addlegendentry{Wiki (PIE)}
  \addplot[smooth, mark=triangle*, dashed, line width=0.6pt, color=blueBelizeHole] coordinates {
  (0.1, 07.335) (0.2, 13.415) (0.3, 18.655) (0.4, 22.48) (0.5, 25.065) (0.6, 28.38) (0.7, 29.44) (0.8, 31.165) (0.9, 32.605) (1, 33.795)
  };\addlegendentry{CWEB-G (PSEUDO)}
  \addplot[smooth, mark=triangle, line width=0.6pt, color=redAlizarin] coordinates { (0.1, 4.39) (0.2, 8.44) (0.3, 12.075) (0.4, 15.45) (0.5, 17.555) (0.6, 18.435) (0.7, 20.575) (0.8, 21.83) (0.9, 23.34) (1.0, 24.41) };\addlegendentry{CWEB-G (PIE)}
\end{axis}
\end{tikzpicture}
    \caption{Precision as a function of the proportion of erroneous sentences in 3 different domains; comparing the \pseudodata (PSEUDO) and PIE systems.}
    \label{fig-percentage}
\end{figure}
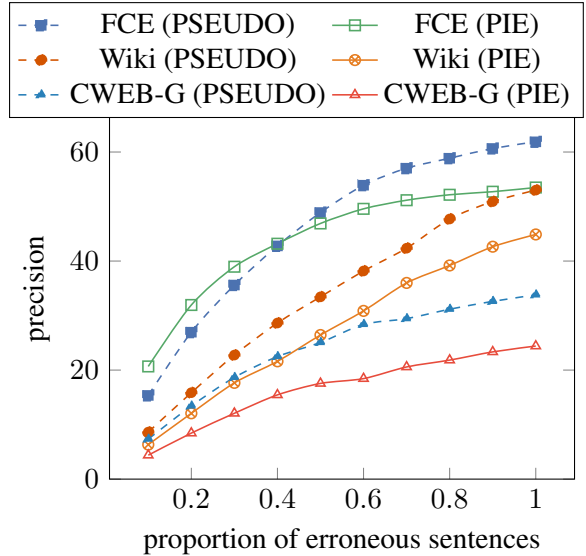

\subsubsection{Analysis of Gold Edits}

\begin{figure*}[!h]

\begin{tikzpicture}

\protected\def\ToSouth#1{%
    \sbox0{#1}%
    \smash{%
      \rlap{%
        \kern-.5\dimexpr\wd0 +\ht0\relax\copy0 %
      }%
    }%
    \hphantom{#1}%
  }
  
\begin{axis}[enlargelimits=0.1, xlabel={Perplexity Ratio}, ylabel={Semantic Similarity}, width=15cm, height=9cm, xtick distance=0.05]
    \addplot[
        scatter/classes={a={black}, b={black}},
        scatter, mark=*, only marks, 
        scatter src=explicit symbolic,
        nodes near coords*={\Label},
        visualization depends on={value \thisrow{label} \as \Label} 
    ] table [meta=class] {
        x y class label

0.8323449523600088 0.9771595556354665 a CoNLL14
0.8313768372468446 0.9922982935632796 a CWEB-G
0.8974376207967738 0.9934783695842642 a CWEB-S
0.844045528413201 0.9883412936294474 a GMEG-Wiki
0.7653876478791306 0.987468170446075 a \ToSouth{GMEG-Yahoo}
0.6919536486324057 0.9768255725675676 a W\&I-A
0.7683817343627034 0.9757823674106577 a W\&I-B
0.8214712224345427 0.9896968405315615 a W\&I-C
0.782805972355341 0.9874421695221837 a LOCNESS
0.6995746067097188 0.97888242575076 a JFLEG
0.7301657485036478 0.9748986589754687 a FCE
0.8822150403260832 0.9931793516692119 a \ToSouth{AESW}
    };
\end{axis}
\end{tikzpicture}
\caption{Average semantic similarity and perplexity ratio (sentence improvement) of sentences before and after being edited, plotted per dataset. The analysis is limited to sentences containing exactly one edit.} 
\label{fig-edit-analysis}
\end{figure*}
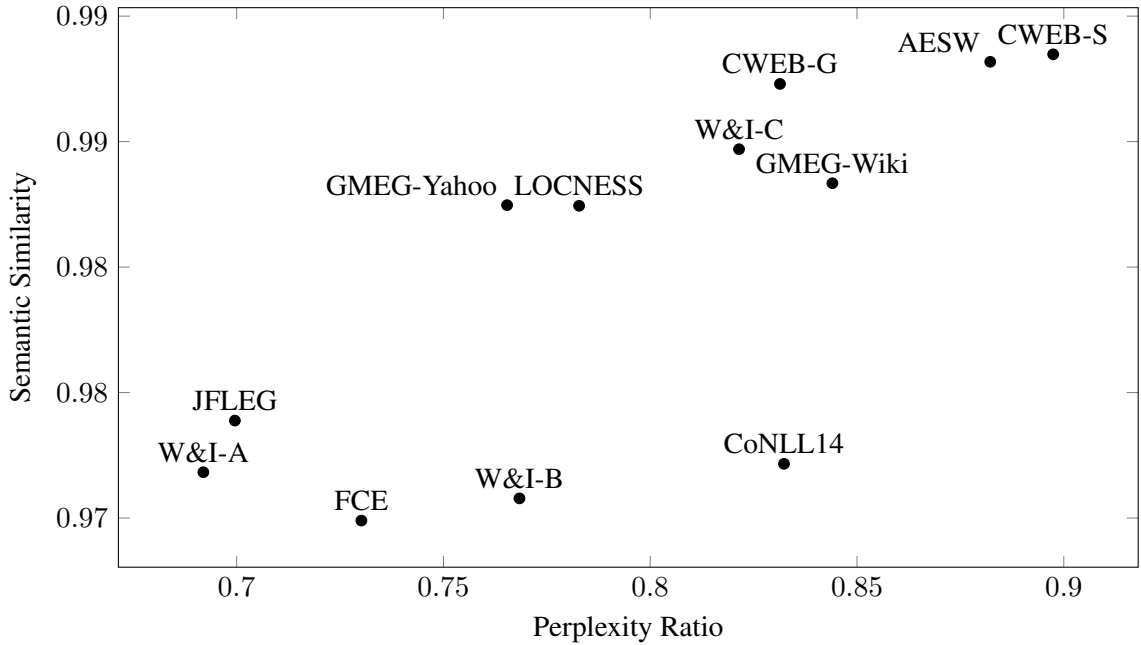
In addition to error density, the type of errors present in the dataset also has an impact on the performance of GEC systems. We investigate how errors and their corresponding corrections differ across domains. In particular, we look at how gold edits in different domains change the sentence in terms of two factors:
1) How much do edits change the semantics of the sentence, and 2) to what degree do edits improve the sentence.

We limit our analysis to sentences containing exactly one edit, as we are interested in how individual edits change a sentence, regardless of how domains differ in amounts of erroneous sentences and in the number of edits per sentence (Table \ref{tab-stats}).

Regarding 1), to measure the semantic change of a sentence 
after an edit is introduced, we use sentence embeddings generated by Sentence-BERT \citep{devlin2018bert} and calculate the cosine similarity between the original sentence and its corrected counterpart. 
Regarding 2), the degree of sentence improvement is calculated as the ratio of the perplexity of GPT-2 \citep{Radford2019LanguageMA} on a sentence after and before it has been edited. 
$$  \Delta P =  \frac{PPL(edited\_sentence)}{PPL (original\_sentence)}$$
A lower ratio suggests that the edited sentence is an improvement, since its perplexity is lower than the original sentence.

Using the outputs of machine learning models as a proxy for semantic change and sentence improvement inevitably introduces biases, but nevertheless provide valuable insights into domain differences.

\paragraph{Corpus Level}
In Figure~\ref{fig-edit-analysis}, the average semantic similarity and perplexity ratio is plotted for each dataset. It is evident that ESL datasets consist of edits with a higher degree of semantic change and sentence improvements than datasets from more advanced speakers. CWEB and AESW in particular stand out, with edits that largely retain the semantics of a sentence and that result in more subtle improvements.

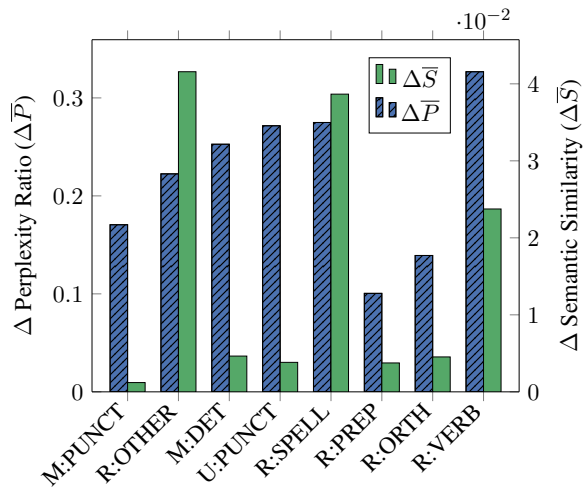
\begin{figure}[!t]
\scalebox{0.82}{
\hspace{-0.3cm}
\begin{tikzpicture}[
   declare function={
    barW=8pt; 
    barShift=barW/2; 
  }
]
\begin{axis}[
    ybar,
    bar width=barW, 
    bar shift=-barShift, 
    symbolic x coords={M:PUNCT, R:OTHER, M:DET, U:PUNCT, R:SPELL, R:PREP, R:ORTH, R:VERB},
    axis y line*=left,
    axis x line=none,
    ymin=0, 
    ylabel=$\Delta$ Perplexity Ratio ($\Delta{\overline{P}}$),
    enlarge x limits=0.1,
    xtick={M:PUNCT, R:OTHER, M:DET, U:PUNCT, R:SPELL, R:PREP, R:ORTH, R:VERB},
    x tick label style={rotate=45, anchor=east},
 ]
    \addplot[
             mark options={xshift=-barShift}, 
             draw=black,
             fill=seabornBlue,
             postaction={pattern=north east lines},
             error bars/.cd,
               y dir=both,
               y explicit,
               error bar style={line width=1pt,solid, black}
    ] coordinates {
    (M:PUNCT, 0.1705503206036877)
    (R:OTHER, 0.22249211469274477)
    (M:DET, 0.2527181985606701)
    (R:SPELL, 0.27491683686937585)
    (U:PUNCT, 0.2715743869692868)
    (R:PREP, 0.10051276874283932)
    (R:ORTH, 0.1391779567895951)
    (R:VERB, 0.32671657574985535)
    }; \label{plot_one}
\end{axis} 
\begin{axis}[
    ybar,
    bar width=barW,
    bar shift=barShift,
    symbolic x coords={M:PUNCT, R:OTHER, M:DET, U:PUNCT, R:SPELL, R:PREP, R:ORTH, R:VERB},
    axis y line*=right,
    ymin=0, 
    ylabel=$\Delta$ Semantic Similarity ($\Delta{\overline{S}}$),
    enlarge x limits=0.1,
    xtick={M:PUNCT, R:OTHER, M:DET, U:PUNCT, R:SPELL, R:PREP, R:ORTH, R:VERB},
    x tick label style={rotate=45, anchor=east},
    legend style={at={(0.84,0.95)}}
]

\addplot[
         mark options={xshift=barShift}, 
         fill=seabornGreen,
         error bars/.cd,
           y dir=both,
           y explicit,
           error bar style={line width=1pt,solid, black}
] coordinates {
(M:PUNCT, 0.0012039255350693168)
(R:OTHER, 0.04157258932834851)
(M:DET, 0.004645047694633364)
(R:SPELL, 0.03866237301024611)
(U:PUNCT, 0.003829282505602327)
(R:PREP, 0.0037525770884521137)
(R:ORTH, 0.004533936304347952)
(R:VERB, 0.023747930193910305)
    }; \label{plot_two}
    
\addlegendimage{/pgfplots/refstyle=plot_one}\addlegendentry{$\Delta{\overline{S}}$}
\addlegendimage{/pgfplots/refstyle=plot_two}\addlegendentry{$\Delta{\overline{P}}$}

\end{axis}

\end{tikzpicture}
}

\caption{Difference in semantic similarity and perplexity ratio between CWEB-S and FCE for the most frequent error types (M: missing; R: replace; U: unnecessary). }
\label{fig-edit-analysis-per-err-type}
\end{figure}

\paragraph{Error type level}
In order to gain further insight on what is driving the differences between datasets, we look separately at how edits of each error type change the sentence.
We compare FCE and CWEB-S, which lie at opposite ends in Figure \ref{fig-edit-analysis}. 
For each dataset, we obtain an average of semantic similarity, $\overline{S}$, and perplexity ratio, $\overline{P}$, separately for sentences of each error type. Then, for each error type, the difference, $\Delta$, between scores in the two datasets is calculated.
$$
 \Delta{\overline{S}} = \overline{S}_{\textrm{CWEB-S}}-\overline{S}_{\textrm{FCE}} \nonumber
$$
$$
\Delta{\overline{P}} = \overline{P}_{\textrm{CWEB-S}}-\overline{P}_{\textrm{FCE}}  \nonumber    
$$
Figure \ref{fig-edit-analysis-per-err-type} plots these differences for the most common error types. We can observe that, for all error types, edits in CWEB-S result in both a lower degree of semantic change and sentence improvement than edits in FCE. 
This is particularly evident for the error types R:OTHER, R:SPELL and R:VERB. These are open class errors, where the error and correction can be quite different. It is therefore reasonable that differences in edits' degree of semantic change and perplexity improvement across domains are particularly observed in these cases.\footnote{Score differences for the R:SPELL error type seem to be driven by a different propensity of spelling errors being of a typographical vs.~phonetical nature in the two datasets.
}

\begin{table}[t]
    \centering
    \begin{small}
    \begin{tabular}{lrrr}
        \toprule
        & \textbf{P} & \textbf{R} & \textbf{\fhalf}  \\
        
\midrule JFLEG & 57.55 & 21.59 & 43.07 \\
\midrule FCE & 51.33 & 17.39 & 36.92 \\
\midrule CoNLL14 & 40.30 & 16.56 & 31.17 \\
\midrule W\&I-A & 45.79 & 15.10 & 32.55 \\ 
W\&I-B & 43.17 & 14.46 & 30.90 \\
W\&I-C & 33.02 & 9.81 & 22.42 \\
\midrule LOCNESS & 42.09 & 16.09 & 31.81 \\
\midrule GMEG Wiki & 52.36 & 13.35 & 32.99 \\
GMEG Yahoo & 62.50 & 16.45 & 39.45 \\
\midrule AESW & 10.18 & 3.58 & 7.44 \\
\midrule CWEB-G & 15.20 & 5.96 & 11.54 \\
CWEB-S & 8.94 & 1.33 & 4.17 \\
        \bottomrule

    \end{tabular}
    \end{small}
    \caption{Scores of a language model based GEC system. 
    The lower scores on CWEB and AESW indicate an inability to rely on language modelling in low error-density domains.
    }
    \label{tab-lmgec}
\end{table}

\subsection{Language Model Importance}

We also investigate the degree to which systems can rely on a strong internal language model representation when evaluated against different domains. 
We examine this by looking at the performance of a purely language
model based GEC system over the different datasets.

We build on the approach of \citet{bryant-briscoe-2018-language}, using confusion sets to generate alternative versions of an input sentence and then deciding if any of the alternatives are preferable to the original version, based on language model probabilities. The authors use an n-gram  language model, which we replace with \mbox{GPT-2} \citep{Radford2019LanguageMA} to see how a strong neural language model performs -- this approach is similar to \citet{Alikaniotis2019TheUE}. 
Hyperparameters are tuned for each dataset (see Appendix \ref{app-lmgec_params} for details).

Table \ref{tab-lmgec} displays the results on the different datasets. Recall and, in particular, precision is substantially lower on CWEB and AESW compared to other datasets. In general, scores are higher in domains with a higher proportion of errors and those containing edits which result in high perplexity improvements. 
In these cases systems can rely on a rough heuristic of replacing low probability sequences with high probability ones. However, in CWEB, where errors are fewer and more subtle, this leads to low precision, as perplexity alone cannot differentiate an erroneous sequence from a sequence that is rare but correct. Table \ref{tab-fp-examples} displays several examples of this, where false positive corrections suggested by the language model based GEC system have large perplexity improvements.

This analysis suggests that the inability to rely on a strong internal language model representation can negatively impact SOTA system performance on CWEB and on low error density domains in general. 
This would mean that having large amounts of error examples for training is more important in high-level domains.

\begin{table}[t]
    \centering
    \begin{small}
    \begin{tabular}{p{0.28\textwidth}|c}
        \toprule
        \textbf{False Positive Examples} & \textbf{Perplexity ratio} \\
        \midrule
        
        All types of work are \sout{callings} \textcolor{red}{\textbf{called}} to individuals. &  0.34  \\
        Get started \sout{at} \textcolor{red}{\textbf{with}} ACC &  0.51  \\
        That \sout{is} \textcolor{red}{\textbf{was}} actually kind of fun! & 0.69 \\
        

        \bottomrule
    \end{tabular}
    \end{small}
    \caption{Examples of false positives on the CWEB dataset that improve perplexity substantially -- even more than the average gold edit in CWEB ($0.86$ perplexity ratio).
    }
    \label{tab-fp-examples}
\end{table}

\section{Conclusion}
We release a new GEC benchmark, CWEB, consisting of  website text generated by English speakers at varying levels of proficiency.  
Comparisons against existing benchmarks demonstrate that CWEB differs in many respects: 1) in the distribution of sentences (higher vocabulary variation and named entity frequency); 2) in error density (lower); and 3) in the types of edits and their impact on language model perplexity and semantic change. 

We showed that existing state-of-the-art GEC models achieve considerably lower performance when evaluated on this new domain, even after fine-tuning. 
We argue that a factor behind this is the inability of systems to rely on a strong internal language model in low error density domains. 

We hope that the dataset shall broaden the target domain of GEC beyond learner and/or exam writing and facilitate the development of robust GEC models in the open-domain setting.

\bibliographystyle{acl_natbib}
\bibliography{references}

\vfill\clearpage
\onecolumn
\appendix

\section{Dataset Download Links}
\label{app-datasets}
\begin{itemize}
    \item JFLEG: \url{https://github.com/keisks/jfleg}
    \item FCE: \url{https://www.cl.cam.ac.uk/research/nl/bea2019st/#data}
    \item CoNLL14: \url{https://www.comp.nus.edu.sg/~nlp/conll14st.html}
    \item Write\&Improve-A/B/C: \url{https://www.cl.cam.ac.uk/research/nl/bea2019st/#data}
    \item LOCNESS: \url{https://www.cl.cam.ac.uk/research/nl/bea2019st/#data}
    \item GMEG Yahoo/Wiki: \url{https://github.com/grammarly/GMEG}
    \item AESW: \url{http://textmining.lt/aesw/aesw2016down.html}
\end{itemize}

\section{Non-averaged Fine-tuning Scores}
\label{non-averaged-scores}

\begin{table}[h]
    \centering
    \begin{tabular}{lrcc}
        \toprule
        & \textbf{P} & \textbf{R} & \textbf{\fhalf}  \\
            \midrule CWEB-G & 53.88 & 34.24 & 48.33 \\
            CWEB-S & 43.65 & 31.1 & 40.39 \\
            \textbf{{CWEB}} (all) & 50.25 & 33.2 & 45.57 \\
        \bottomrule
    \end{tabular}
    \caption{Scores of the \pseudodata system fine-tuned on CWEB data, calculated against both annotators.
    }
    \label{tab-finetuning-non-averaged}
\end{table}

\section{Language Model GEC Hyperparameter Tuning}
\label{app-lmgec_params}
 
A threshold, $\tau$, determines the degree of probability improvement needed before an alternative sentence is preferred. 
For each dataset, we find $\tau$, in the $0.9$ to $1.0$ range, resulting in the best development set $F_{0.5}$. For CoNLL14, we tune on CoNLL13; for W\&I, we use the dedicated training sets; for LOCNESS, there is no training set available and so we tune on the W\&I subset of advanced texts (W\&I-C).

\begin{table*}[h]
    \centering
    \begin{small}
    \begin{tabular}{ll@{~~}r@{~}r@{~~~}rrr@{~~~}r@{~~~}r@{~~~}r@{~~~~}r@{~~~}rr@{~~~}r}
        \toprule & \multirow{2}{*}{\textbf{JFLEG}} & \multirow{2}{*}{\textbf{FCE}} & \multirow{2}{*}{\textbf{CoNLL14}} & \multicolumn{3}{c}{\textbf{W\&I}} & \multirow{2}{*}{\textbf{LOCNESS}} &  \multicolumn{2}{c}{\textbf{GMEG}} & 
        \multirow{2}{*}{\textbf{AESW}} &  \multicolumn{3}{c}{\textbf{CWEB}} \\
         \cmidrule(r){5-7} \cmidrule(r){9-10} \cmidrule(r){12-14} 
        & & & & \textbf{A} & \textbf{B} & \textbf{C} & & \textbf{Wiki} & \textbf{Yahoo} & &  \textbf{G} & \textbf{S} & \\
    
        \midrule
        $\tau$ & 0.97 & 0.97 & 0.98 & 0.98 & 0.98 & 0.97 & 0.97 & 0.96 & 0.91 & 0.96 & 0.96 & 0.93 &  \\
 
        \bottomrule
    \end{tabular}
    \end{small}
    \caption{Best performing threshold $\tau$ for each domain.}

\end{table*}

\section{Precision as a Function of the Proportion of Erroneous Sentences}
\label{app-percentage}

\begin{figure*}[h]
\begin{tikzpicture}
  \begin{groupplot}[width=\linewidth/2,
  group style = {group size = 2 by 2, vertical sep=90, horizontal sep=30},
  ymax=65,
  ymin=0,
  title style = {yshift=.02cm,xshift=-4cm},
  ylabel style = {yshift=0},
  legend style = {at={(1,-.15)}, legend columns=3},
  ]
  \nextgroupplot[ylabel={precision}]
  \addplot[smooth, mark=square*, dotted, line width=0.6pt, color=seabornBlue] coordinates {
  (0.1, 15.35) (0.2, 26.95) (0.3, 35.6) (0.4, 42.78) (0.5, 48.9) (0.6, 53.91) (0.7, 57.03) (0.8, 58.86) (0.9, 60.65) (1.0, 61.87)
  }; 
  \addplot[smooth, mark=triangle*, dotted, line width=0.6pt, color=seabornGreen] coordinates { 
  (0.1, 16.42) (0.2, 27.515) (0.3, 34.677) (0.4, 40.3) (0.5, 44.483) (0.6, 48.295) (0.7, 50.707) (0.8, 52.67) (0.9, 54.955) (1.0, 56.203)
  }; 
  \addplot[smooth, mark=triangle, line width=0.6pt, color=blueBelizeHole] coordinates {
  (0.1, 7.65) (0.2, 14.732) (0.3, 19.907) (0.4, 25.305) (0.5, 29.852) (0.6, 33.903) (0.7, 37.333) (0.8, 39.46) (0.9, 41.068) (1.0, 42.11)
  }; 
  \addplot[smooth, mark=square, line width=0.6pt, color=orangePumpkin] coordinates {
  (0.1, 8.62) (0.2, 15.855) (0.3, 22.755) (0.4, 28.635) (0.5, 33.435) (0.6, 38.18) (0.7, 42.357) (0.8, 47.715) (0.9, 50.907) (1.0, 53.012)
  }; 
  \addplot[smooth, mark=*, dashed, line width=0.6pt, color=orangeCarrot] coordinates {
  (0.1, 5.725) (0.2, 10.535) (0.3, 14.695) (0.4, 18.13) (0.5, 21.575) (0.6, 24.845) (0.7, 27.75) (0.8, 29.61) (0.9, 30.8749) (1.0, 32.58)
  }; 
  \addplot[smooth, mark=triangle*, dashed, line width=0.6pt, color=redAlizarin] coordinates {
  (0.1, 07.335) (0.2, 13.415) (0.3, 18.655) (0.4, 22.48) (0.5, 25.065) (0.6, 28.38) (0.7, 29.44) (0.8, 31.165) (0.9, 32.605) (1, 33.795)
  }; 
  \legend{FCE~2.1,JFLEG, Yahoo, Wiki, CWEB-S, CWEB-G}
  \nextgroupplot[title={\pseudodata system}]
  \addplot[smooth, mark=square*, dotted, line width=0.6pt, color=seabornBlue] coordinates { 
  (0.1, 22.38) (0.2, 34.24) (0.3, 42.19) (0.4, 45.61) (0.5, 51.23) (0.6, 54.65) (0.7, 56.28) (0.8, 58.3) (0.9, 59.66) (1.0, 60.68)
  }; 
  \addplot[smooth, mark=triangle*, dotted, line width=0.6pt, color=seabornGreen] coordinates { 
  (0.1, 20.55) (0.2, 33.25) (0.3, 41.98) (0.4, 46.84) (0.5, 51.0) (0.6, 54.28) (0.7, 57.51) (0.8, 60.14) (0.9, 61.27) (1.0, 62.69)
  }; 
  \addplot[smooth, mark=*, loosely dashed, line width=0.6pt, color=blueBelizeHole] coordinates { 
  (0.1, 15.51) (0.2, 26.04) (0.3, 34.26) (0.4, 40.02) (0.5, 44.85) (0.6, 48.37) (0.7, 50.85) (0.8, 51.81) (0.9, 54.75) (1.0, 56.58)
  }; 
  \addplot[smooth, mark=diamond*, loosely dashed, line width=0.6pt, color=orangePumpkin] coordinates { 
  (0.1, 18.1) (0.2, 30.59) (0.3, 40.62) (0.4, 45.78) (0.5, 50.52) (0.6, 53.86) (0.7, 56.72) (0.8, 58.16) (0.9, 59.69) (1.0, 61.51)
  }; 
  \addplot[smooth, mark=*, dotted, line width=0.6pt, color=orangeCarrot] coordinates { 
  (0.1, 11.555) (0.2, 19.405) (0.3, 26.68) (0.4, 32.55) (0.5, 36.25) (0.6, 39.29) (0.7, 42.22) (0.8, 44.865) (0.9, 46.71) (1.0, 47.725)
  }; 
  \addplot[smooth, mark=otimes, line width=0.6pt, color=redAlizarin] coordinates {
  (0.1, 4.42) (0.2, 8.38) (0.3, 12.04) (0.4, 15.27) (0.5, 18.17) (0.6, 20.9) (0.7, 23.08) (0.8, 25.58) (0.9, 27.52) (1.0, 29.37)
  }; 
  ;\legend{W\&I-A,W\&I-B,W\&I-C,LOCNESS, CONLL14, AESW}
  \nextgroupplot[ylabel={precision},xlabel={proportion of erroneous sentences}]
  \addplot[smooth, mark=square*, dotted, line width=0.6pt, color=seabornBlue] coordinates { 
  (0.1, 20.67) (0.2, 31.94) (0.3, 38.96) (0.4, 43.18) (0.5, 46.9) (0.6, 49.58) (0.7, 51.16) (0.8, 52.16) (0.9, 52.72) (1.0, 53.49)
  }; 
  \addplot[smooth, mark=triangle*, dotted, line width=0.6pt, color=seabornGreen] coordinates { 
  (0.1, 18.83) (0.2, 28.252) (0.3, 36.162) (0.4, 38.693) (0.5, 43.275) (0.6, 46.705) (0.7, 47.485) (0.8, 48.995) (0.9, 50.0) (1.0, 51.075)
  }; 
  \addplot[smooth, mark=triangle, line width=0.6pt, color=blueBelizeHole] coordinates {
  (0.1, 6.195) (0.2, 12.218) (0.3, 16.77) (0.4, 21.438) (0.5, 25.61) (0.6, 28.633) (0.7, 32.77) (0.8, 34.0) (0.9, 34.782) (1.0, 35.37)
  }; 
  \addplot[smooth, mark=square, line width=0.6pt, color=orangePumpkin] coordinates {
  (0.1, 6.325) (0.2, 12.07) (0.3, 17.655) (0.4, 21.605) (0.5, 26.433) (0.6, 30.857) (0.7, 35.987) (0.8, 39.188) (0.9, 42.633) (1.0, 44.877)
  }; 
  \addplot[smooth, mark=*, dashed, line width=0.6pt, color=orangeCarrot] coordinates {
  (0.1, 1.65) (0.2, 3.25) (0.3, 4.89) (0.4, 6.215) (0.5, 7.695) (0.6, 8.745) (0.7, 9.56) (0.8, 9.935) (0.9, 11.35) (1.0, 12.19)
  }; 
  \addplot[smooth, mark=triangle*, dashed, line width=0.6pt, color=redAlizarin] coordinates {
  (0.1, 4.39) (0.2, 8.44) (0.3, 12.075) (0.4, 15.45) (0.5, 17.555) (0.6, 18.435) (0.7, 20.575) (0.8, 21.83) (0.9, 23.34) (1.0, 24.41)
  }; 
  \nextgroupplot[title={PIE system},xlabel={proportion of erroneous sentences}]
  \addplot[smooth, mark=square*, dotted, line width=0.6pt, color=seabornBlue] coordinates { 
  (0.1, 21.74) (0.2, 32.9) (0.3, 40.35) (0.4, 43.74) (0.5, 46.64) (0.6, 49.46) (0.7, 51.66) (0.8, 52.8) (0.9, 53.8) (1.0, 54.25)
  }; 
  \addplot[smooth, mark=triangle*, dotted, line width=0.6pt, color=seabornGreen] coordinates { 
  (0.1, 18.75) (0.2, 28.6) (0.3, 36.43) (0.4, 40.52) (0.5, 45.94) (0.6, 47.7) (0.7, 50.16) (0.8, 50.96) (0.9, 52.23) (1.0, 52.61)
  }; 
  \addplot[smooth, mark=*, loosely dashed, line width=0.6pt, color=blueBelizeHole] coordinates { 
  (0.1, 12.99) (0.2, 22.29) (0.3, 30.91) (0.4, 34.87) (0.5, 38.97) (0.6, 42.0) (0.7, 43.64) (0.8, 44.54) (0.9, 46.32) (1.0, 47.34)
  }; 
  \addplot[smooth, mark=diamond*, loosely dashed, line width=0.6pt, color=orangePumpkin] coordinates { 
  (0.1, 12.34) (0.2, 20.8) (0.3, 27.4) (0.4, 31.58) (0.5, 35.51) (0.6, 38.24) (0.7, 40.28) (0.8, 41.79) (0.9, 43.17) (1.0, 44.07)
  }; 
  \addplot[smooth, mark=*, dotted, line width=0.6pt, color=orangeCarrot] coordinates {
  (0.1, 12.6) (0.2, 21.205) (0.3, 27.25) (0.4, 32.65) (0.5, 35.905) (0.6, 39.44) (0.7, 41.705) (0.8, 43.74) (0.9, 45.525) (1.0, 46.015)
  }; 
  \addplot[smooth, mark=otimes, line width=0.6pt, color=redAlizarin] coordinates {
  (0.1, 02.63) (0.2, 5.1) (0.3, 7.44) (0.4, 9.43) (0.5, 10.96) (0.6, 12.37) (0.7, 13.62) (0.8, 14.65) (0.9, 15.67) (1.0, 16.49)
  }; 
  \end{groupplot}
\end{tikzpicture}
 \caption*{Precision as a function of the proportion of erroneous sentences in each domain.}\label{fig-percentage-all}
\end{figure*}
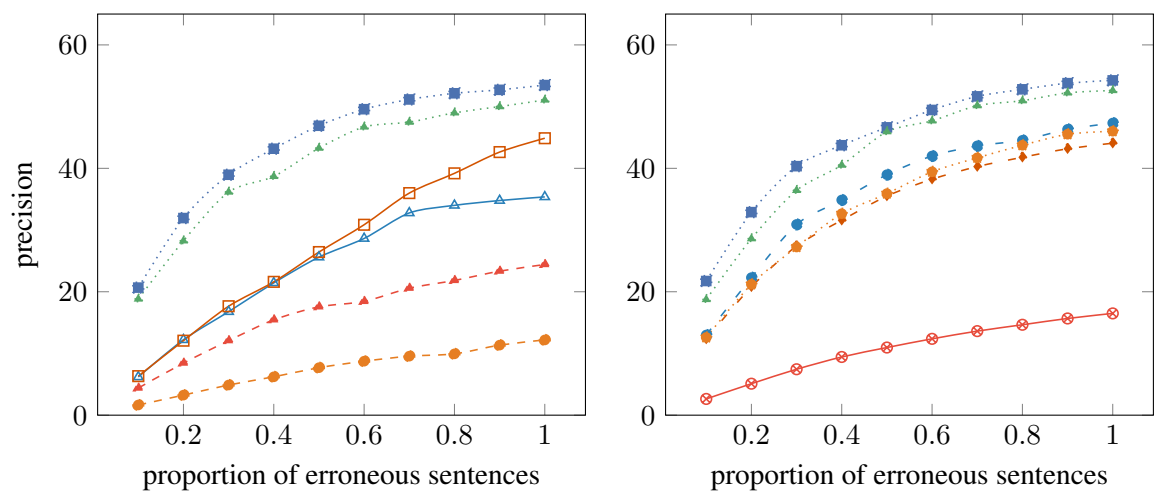

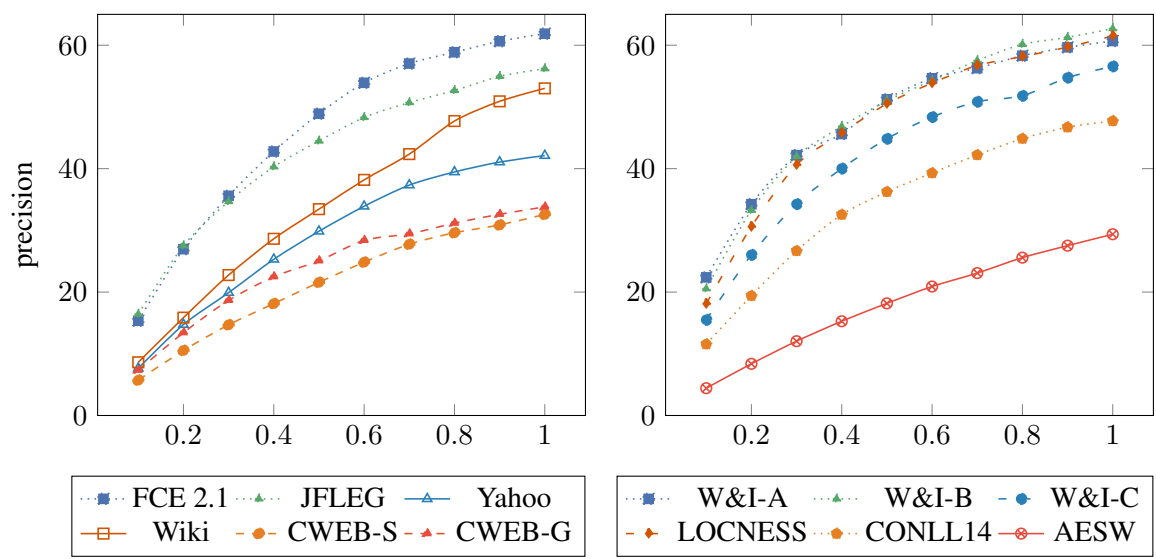
\end{document}